\documentclass[10pt]{cai26}

\begin{document}
\def\conferenceyear{2026}
\volumeheader{39}{0}
\begin{center}

\title{Fed-Universe: A Semantic-Geometric-Topological-Human (S-G-T-H) Stack for Negotiated Alignment in Federated Systems}
\maketitle

\thispagestyle{empty}
\pagenumbering{gobble}

\begin{tabular}{cc}
Ting Xu\upstairs{\affilone,*} \quad Henry Leung\upstairs{\affiltwo}
\\[0.25ex]
{\small \upstairs{\affilone} Department of Biomedical Engineering, University of Calgary, Calgary, Canada} \\
{\small \upstairs{\affiltwo} Department of Electrical and Software Engineering, University of Calgary, Calgary, Canada} \\[0.5ex]
\end{tabular}

\emails{
  \upstairs{*} tina.xu1@ucalgary.ca
}
\vspace*{0.1in}
\end{center}

\begin{abstract}
Conventional federated learners implicitly optimize for global means, creating a ``one-size-fits-all'' paradigm that inevitably suppresses the minority under cross-site heterogeneity. To bridge this critical gap, we present \textit{Fed-Universe}, a generalizable Semantic--Geometric--Topological--Human ($\mathcal{S}$--$\mathcal{G}$--$\mathcal{T}$--$\mathcal{H}$) architecture that transforms passive averaging into an active negotiation for decentralized alignment. The $\mathcal{S}$-layer deploys an edge-capable LLM as a semantic surrogate to translate heterogeneous inputs into standardized patient profile prompts. The $\mathcal{G}$-layer enforces geometric quality assurance (GQA) via a mapping $f: \Theta \to Q$ from the statistical parameter space to a physically interpretable quality domain, employing a cosine similarity gate to retain minority updates based on physical geometry rather than statistical proximity. The $\mathcal{T}$-layer executes topological Pareto control, identifying a sensitivity-based knee point ($\lambda_k$) to balance multi-objective aggregation without majority degradation. Finally, the $\mathcal{H}$-layer operationalizes a cognitive twin dynamic to mirror the user's real-time psychological state, dynamically adapting human-computer interaction modes via active intent alignment. To ground this theoretical architecture empirically, we validated the $\mathcal{S}$-$\mathcal{G}$-$\mathcal{T}$ core on a binational synthetic clinical network simulation. This proof-of-concept demonstrated a ``zero-sum escape'': a minority node ($<$3\% data volume) achieved utility parity (loss reduction from 0.857 to 0.340) alongside dominant hubs. Our \textit{Fed-Universe} framework is being expanded along two focused pillars: institutional multimodal alignment (\textit{Fed-Ultra}) and individual cognitive intent alignment (\textit{Fed-Human}).
\end{abstract}

\copyrightnotice


\section{Introduction}
Standard federated learning paradigms (e.g., FedAvg \cite{mcmahan2017communication}) operate under a ``one-size-fits-all'' objective. By implicitly optimizing for the global mean, they act as passive systemic averagers \cite{li2020federated, yang2019federated}. In non-IID real-world settings, the global model tends to overfit dominant hubs while neglecting local heterogeneity \cite{mohri2019agnostic, kairouz2021advances}, such as cross-site physiological variation, rare patient subpopulations, or human emotional variance. Furthermore, in human-computer interaction (HCI), rigid models often discard cognitive variance as noise. For instance, if an AI treats a user's frustration as noise and resets to a blank slate (the ``stateless user paradox''), the misunderstood user grows more frustrated. This drives the user to abandon the interaction, eventually leading to ``feedback collapse'' where the system also learns nothing from the user \cite{amershi2019guidelines}.

To bridge this critical gap, we propose \textit{Fed-Universe}. Named for its universal adaptability, it serves as a foundational meta-architecture capable of deployment across diverse dimensions: \textbf{Fed-Ultra} (context-aware medical imaging) and \textbf{Fed-Human} (human-aware HCI intent alignment). Unlike personalized federated learning (PFL) that isolates local models, \textit{Fed-Universe} maintains a collaborative knowledge commons by acting as an active negotiator, NOT an averager. It operationalizes a Semantic--Geometric--Topological--Human ($\mathcal{S}$-$\mathcal{G}$-$\mathcal{T}$-$\mathcal{H}$) stack to semantically translate heterogeneous contexts, geometrically gate misaligned updates, topologically balance minority representations \cite{hu2022federated}, and dynamically adapt HCI via cognitive mirroring \cite{shneiderman2020human}. By shifting from rigid aggregation to dynamic negotiation, \textit{Fed-Universe} establishes a symbiotic ecosystem where algorithmic intelligence and human cognition continuously co-evolve.

\section{Methods}
\textit{Fed-Universe} is a robust framework that reframes federated learning as a structured negotiation rather than a naive averaging process. Concretely, the architecture operates across four synergistic layers as shown in Figure 1:

\begin{figure}[!ht]
    \centering
    \includegraphics[width=0.9\linewidth]{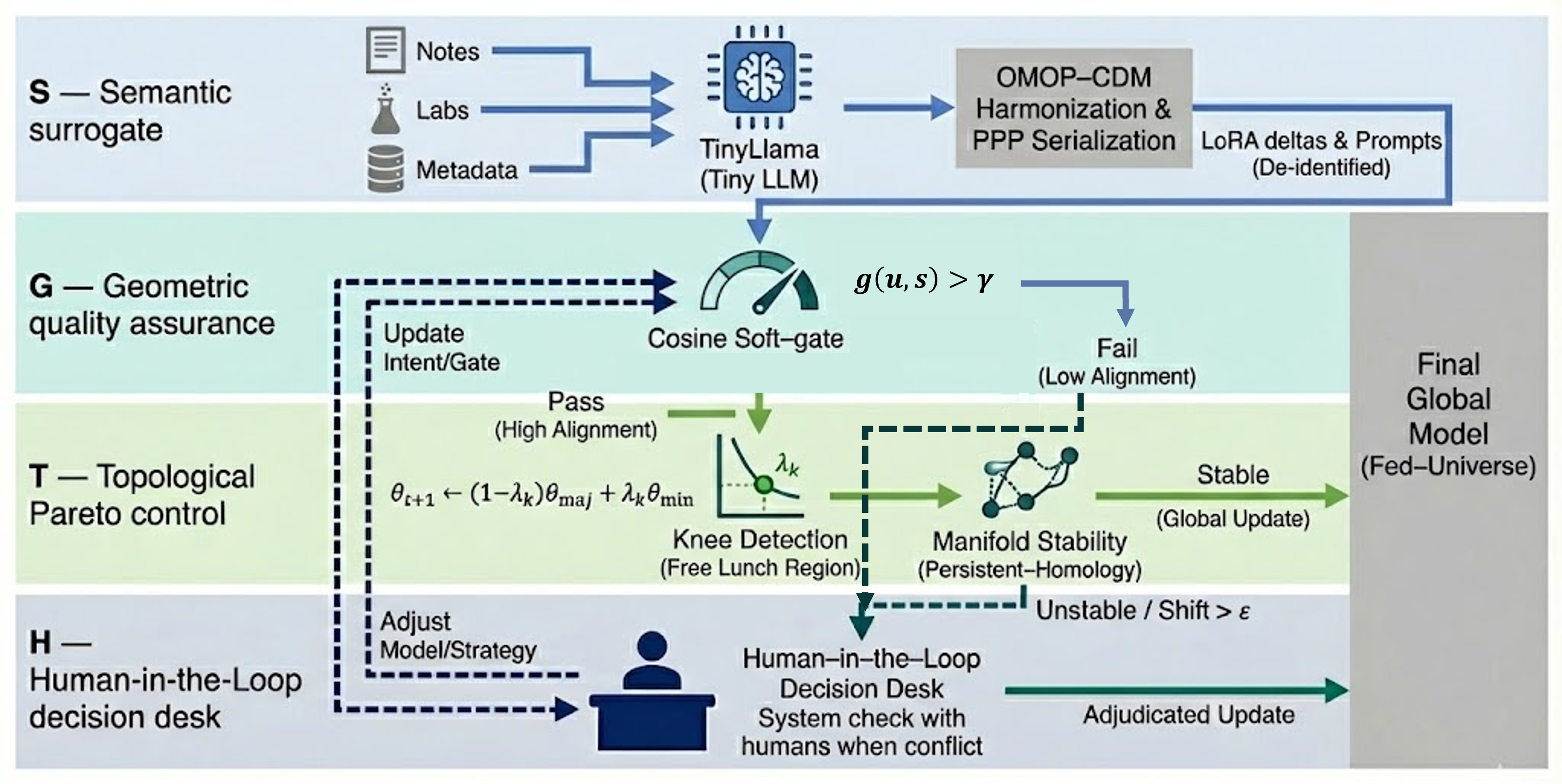}
    \caption{\textbf{The \textit{Fed-Universe} architecture:} A federated $\mathcal{S}$-$\mathcal{G}$-$\mathcal{T}$-$\mathcal{H}$ negotiation stack for decentralized intent alignment and Pareto-optimal fairness, equipped with localized geometric safety gates.}
    \label{fig:architecture}
\end{figure}

\begin{itemize}
    \item \textbf{$\mathcal{S}$-Layer: Semantic surrogate.} A lightweight LLM (e.g., TinyLlama \cite{zhang2024tinyllama}) operates inside each client site as a local data translator. Heterogeneous raw inputs (clinical notes, lab results, metadata) are harmonized to the OMOP-CDM standard \cite{hripcsak2015observational} and serialized into \textbf{patient profile prompts (PPP)}. To ensure stringent privacy and minimize compute overhead, raw files remain strictly localized; only de-identified intents and small LoRA parameter deltas \cite{hu2021lora} are shared with the global server.
    
    \item \textbf{$\mathcal{G}$-Layer: Geometric quality assurance (GQA).} We propose a mapping $f\!: \Theta \to Q$ from the statistical parameter space $\Theta$ to a physically interpretable quality domain $Q$. Conventional distance-based aggregators (e.g., Krum \cite{blanchard2017machine}, RFA \cite{pillutla2022robust}) often conflate valid minority features with adversarial noise. By evaluating directional alignment within $Q$ via a continuous cosine similarity gate:
    $$g(u, s) = \sigma(\beta \cos \theta(u, s) - \tau)$$
    GQA isolates structural validity from statistical frequency. This ensures updates from localized minorities (e.g., high-BMI cohorts) are preserved based on their underlying physical geometry, rather than rejected as anomalies in $\Theta$.

    \item \textbf{$\mathcal{T}$-Layer: Topological Pareto control (TPC).} To systematically balance majority ($\theta_{\text{maj}}$) and minority ($\theta_{\text{min}}$) representations, we formulate the global update as a multi-objective negotiation:
    $$\theta_{t+1} \leftarrow (1 - \lambda_k)\theta_{\text{maj}} + \lambda_k\theta_{\text{min}}$$
    To enforce topological control across the parameter space, we map the empirical Pareto frontier. Specifically, the optimal aggregation coefficient ($\lambda_k$) is algorithmically determined via nonlinear sensitivity analysis. By identifying the point of maximum curvature along the utility trade-off curve (the topological ``knee point''), this approach mathematically targets the ``free lunch region,'' maximizing minority utility without degrading the established majority baseline.
    
    \item \textbf{$\mathcal{H}$-Layer: Human-in-the-Loop (HITL) decision desk.} This layer operationalizes the \textbf{cognitive twin dynamic (CTD)} to mirror the user's real-time context \cite{shneiderman2020human}, enabling the adaptation of localized modes based on captured states. When algorithmic alignment ($g(u, s) < \gamma$) or Pareto stability ($\lambda_k$) fails, the $\mathcal{H}$-layer resolves the resulting ``interaction divergence'' by synchronizing systemic updates with the mirrored human intent. By treating users as dynamic entities rather than static data sources, this alignment prevents feedback collapse and fosters a virtuous cycle of human-AI co-evolution, ultimately ensuring human sovereignty as the failsafe.

\end{itemize}

\section{Results}
We evaluate the $\mathcal{S}$--$\mathcal{G}$--$\mathcal{T}$ pipeline using a binational simulation integrating Canadian and U.S. synthetic clinical registries. The architectural proof-of-concept is explicitly validated: the $\mathcal{S}$-layer standardizes heterogeneous inputs into de-identified LLM prompts, while the $\mathcal{G}$-layer safeguards updates that align with the ``golden direction''. Finally, the $\mathcal{T}$-layer identifies the Pareto-optimal knee point ($w=2.0$, calculated using $\lambda_k \approx 0.67$) to balance the multi-objective aggregation and accommodate underrepresented nodes.

\begin{figure}[!ht]
    \centering
    \includegraphics[width=1.0\linewidth]{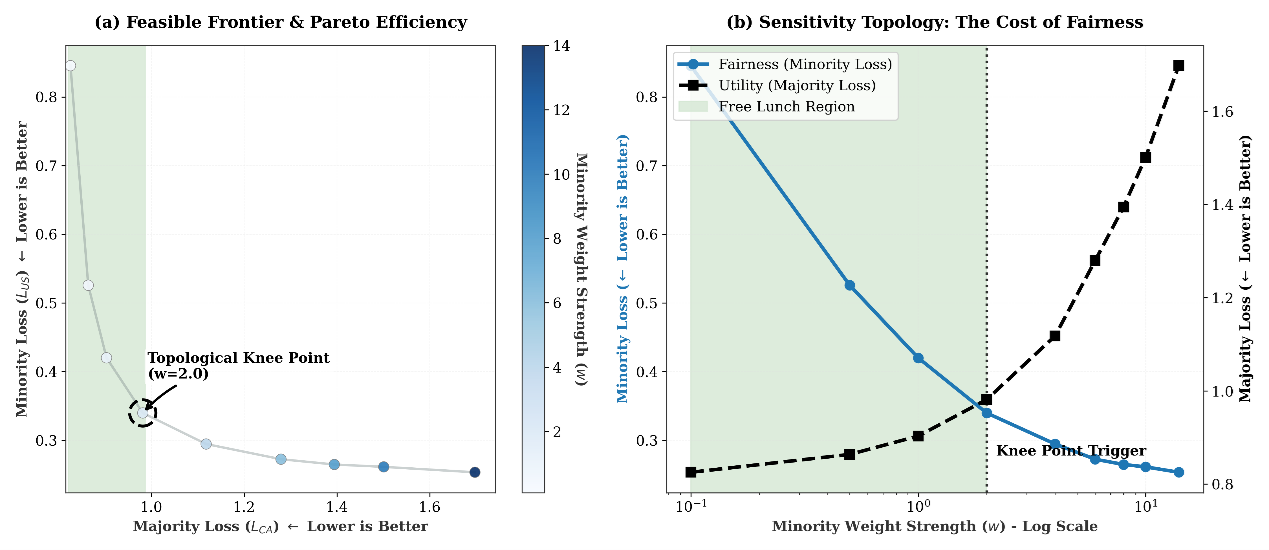}
    \caption{\textbf{Pareto frontier sensitivity analysis:} The identified knee point at $w=2.0$ establishes an optimal equilibrium, maximizing minority spoke utility without degrading the majority hub baseline.}
    \label{fig:pareto}
\end{figure}

As shown in Figure 2, the $\mathcal{S}$--$\mathcal{G}$--$\mathcal{T}$ trajectory deviates from the suboptimal FedAvg equilibrium. Decoupled from volume dominance, the minority node ($<$3\% of total data) achieves utility parity, with its loss decreasing from 0.857 to 0.340. Notably, this gain occurs without degrading the data-rich majority baseline (0.981 vs. 0.985). These findings suggest that the framework effectively captures the ``free-lunch region,'' ensuring that global alignment incorporates rather than suppresses underrepresented patterns. Compared to simple averaging, the pipeline ensures that global generalizability does not come at the cost of local representation, providing the necessary algorithmic foundation for our two focused application pillars: \textit{Fed-Ultra} and \textit{Fed-Human}.

\section{Discussion}
 Having validated the $\mathcal{S}$-$\mathcal{G}$-$\mathcal{T}$ core on a binational synthetic clinical network, the next phase of our thesis is to rigorously test the framework’s generalizability. We structure our framework around two representative pillars of alignment failure in medical AI:
\begin{itemize}
    \item \textbf{Fed-Ultra:} This pillar integrates epidemiological priors (NHANES, SEER) \cite{nguyen2023harmonized} with ultrasound datasets (DDTI, TN3K) via $\mathcal{S}$-layer translation and $\mathcal{T}$-layer fusion. By employing GQA in the $\mathcal{G}$-layer, updates from physiologically distinct cohorts (e.g., high-BMI) are retained based on valid physical structures in $Q$, avoiding mis-classification as statistical anomalies in $\Theta$. We hypothesize this approach will improve early thyroid cancer screening accuracy for underrepresented populations compared to standard statistical gating.

    \item \textbf{Fed-Human:} Addressing human-computer interaction, this pillar mitigates interaction divergence—where model optimization drifts from user needs. The $\mathcal{G}$-layer isolates core user intent from high-entropy emotional variance to provide a reliable alignment vector. Concurrently, the $\mathcal{H}$-layer utilizes cognitive twin dynamics \cite{shneiderman2020human} to adapt system responses based on the user's real-time psychological state, effectively resolving the stateless user paradox \cite{amershi2019guidelines} and establishing a virtuous cycle of adaptive human-AI collaboration.
\end{itemize}
Looking ahead, the $\mathcal{T}$-layer will leverage graph Laplacian regularization as a computationally efficient proxy for mutual information. By embedding client-graph topology directly into the Pareto objective, this formulation aims to minimize edge processing overhead while ensuring robust, deterministic intra-cluster cohesion without relying on heuristic adaptive thresholds.

\section{Conclusion}
\textit{Fed-Universe} reframes federated learning as active negotiation via a four-layer $\mathcal{S}$-$\mathcal{G}$-$\mathcal{T}$-$\mathcal{H}$ stack designed to break the ``one-size-fits-all'' paradigm in federated learning. By transforming passive averaging into structured negotiation, the framework integrates semantic translation ($\mathcal{S}$), geometric quality assurance ($\mathcal{G}$), topological Pareto control ($\mathcal{T}$), and dynamic intent alignment via cognitive twin mirroring ($\mathcal{H}$), accommodating the underrepresented while emphasizing human-centric application. Validated on a binational clinical simulation, the $\mathcal{S}$-$\mathcal{G}$-$\mathcal{T}$ core achieved a ``zero-sum escape'': a minority node ($<$3\% data volume) reduced loss from 0.857 to 0.340 without degrading the majority baseline. These results ground the forthcoming \textit{Fed-Ultra} and \textit{Fed-Human} pillars toward a collaborative, equitable knowledge commons and a robust human-AI partnership.

\section*{Acknowledgements}
This work was supported by the Natural Sciences and Engineering Research Council of Canada (NSERC). The authors gratefully acknowledge the Canadian Network for Observational Drug Effect Studies (CNODES) and the MITRE Corporation (SyntheticMass) for providing the synthetic datasets used in this research. The authors also thank Baihui Chen from Beijing Normal University for her expertise in psychological counseling on the design of the $\mathcal{H}$-layer; her ongoing collaboration is anticipated in the forthcoming \textit{Fed-Human} study.

\section*{Declaration of Generative AI and AI-assisted Technologies}
During the preparation of this work, the author(s) used Large Language Models (including Claude and Gemini) in order to polish the language, improve readability, and assist with \LaTeX{} structural formatting. After using these tools, the author(s) thoroughly reviewed and edited the content as needed and take(s) full responsibility for the final content, scientific accuracy, and integrity of the publication.

\printbibliography[heading=subbibintoc]

\end{document}